\documentclass[a4paper,11pt,final]{article}

\usepackage{epsfig}
\usepackage[cmex10]{amsmath}
\usepackage{amssymb,amsthm}
\usepackage{mathtools}
\usepackage{nccmath}
\usepackage{soul}
\usepackage[normalem]{ulem}
\usepackage{cite}
\usepackage{todonotes}
\usepackage{tikz}
\usepackage{hyperref}
\usetikzlibrary{hobby,decorations.markings,arrows,intersections,shapes}
\usepackage{pgfplots}
\usepgfplotslibrary{fillbetween}
\pgfplotsset{compat=newest}
\usepgfplotslibrary{fillbetween}
\usepackage{verbatim}

\begin{document}

\title{\ \\ \LARGE\bf Do Random and Chaotic Sequences Really Cause Different PSO Performance? Further Results}

\author{Paul Moritz N\"orenberg and Hendrik~Richter \\
HTWK Leipzig University of Applied Sciences \\ Faculty of
Engineering \\
        Postfach 301166, D--04251 Leipzig, Germany. \\ Email:
        moritz.noerenberg@outlook.de\\ 
hendrik.richter@htwk-leipzig.de. }

\maketitle

\begin{abstract}

Empirical results show that PSO performance may be different if using either chaotic or random sequences to drive the algorithm's search dynamics. We analyze the phenomenon by evaluating the performance based on a benchmark of test functions and comparing random and chaotic sequences according to equality or difference in underlying distribution or density. Our results show that the underlying distribution is the main influential factor in performance and thus the assumption of general and systematic performance differences between chaos and random appears not plausible.

\end{abstract}

\section{Introduction}
\label{sec:introduction}
A main driving force of metaheuristic, nature-inspired, population-based optimization methods is random. This particularly applies for particle swarm optimization (PSO). The most common way for obtaining  the needed random sequences
in practical implementations is by employing a pseudo-random number generator (PRNG)~\cite{mat98,lec12}. 
 A popular alternative is using the time evolution of chaotic maps as a substitute for PRNGs. As chaotic dynamics is known to be ergodic, certain maps with appropriate parameters give  sequences with random-like properties comparable to those obtained by PRNGs. Moreover, as chaotic trajectories show a sensitive dependence on initial conditions, different initial states yield different sequences, but for the same map and the same para\-meters they all have the same underlying statistical properties. Thus, they can be seen as equivalent to realizations of a random process.     

During the past decades a multitude of population-based metaheuristic algorithms have been suggested which use sequences from chaotic maps to drive their search dynamics, see for instance~\cite{capo03,chen18,gag21,xu18,ma19,gan13,tian19,plu13,plu14,yang14,zel22}. These works are usually motivated by attempting to identify particularly efficient optimization methods. Thus, analyzing such algorithms at least implicitly implies  discussing  the effect of using either random or chaotic sequences on algorithmic behaviour and performance. Thus, 
differences and similarities in performance between chaotic and random sequences are a topic of extensive  debate in evolutionary computation and we frequently have some claims of chaos showing better performance than random, but different results have also been obtained~\cite{ala09,chen18,gan13,tian19,plu14,zel22,kuang14,rong10,yang12,liu05}.

Recently, a forerunner work~\cite{noer23} contributed to approaches explicitly discussing the effect of chaos on  performance of metaheuristic algorithms~\cite{gag21,zel22}.     Particularly, it focused on  PSO and used a novel approach for asking to what extent random and chaotic sequences really cause different PSO performance. In addition to comparing the performance based on a benchmark of test functions, random and chaotic sequences have been analyzed according to equality or difference in underlying distribution or density. Such an approach allows differentiating between the influence
of the underlying distribution and the effects connected to how the
sequences are generated (PRNGs or chaotic maps). In this paper, we present further results using this approach. In addition to comparing bathtub-shaped distributions (Logistic map and a certain Beta distribution), we also compare distributions of chaotic and random sequences which are bell-shaped (Weierstrass map and normal distribution) and constant (Tent map and uniform distribution). Our results confirm and verify that the underlying distribution is the main  influential factor in performance and the origin of the  sequences (PRNGs or chaotic maps) is secondary.



The paper is structured as follows. In Section 2  we briefly review generating random and chaotic sequences, introduce the chaotic maps used and compare their short-term and long-term properties such as  invariant and probability densities as well as the decay of the autocorrelation. The PSO algorithm and the experimental setup are discussed in Section 3. The results are presented in Section 4 and we close the paper with a discussion of the results and concluding remarks.

\section{Random and chaotic sequences} \label{sec:rand_chao}
Consider the Logistic map
\begin{equation}z(k+1)=rz(k)(1-z(k)) \label{eq:logist} \end{equation}
defined on the (state space) interval $[0,1]$. 
For certain parameter values of $r$ and  initial states $z(0)\in\mathbb{R}$ it yields  a chaotic trajectory $z=(z(0),z(1),\ldots,z(k),z(k+1))$ which can be interpreted as a sequence of real numbers. 
Statistical properties of chaotic sequences can be described by the natural invariant density~\cite{ott93,diak96}. The invariant density expresses the distribution of values over the chaotic map's state space, which is equivalent to a probability distribution for random sequences over a sample space.
For $r=4$ and  $k \rightarrow \infty$, the sequence has the
natural invariant density, see~\cite{ott93}, p. 33
\begin{equation}\varrho(z)=\left(\pi \sqrt{z(1-z))}\right)^{-1}. \label{eq:log_dens} \end{equation}
Thus, a chaotic sequence $z$ of the Logistic map (\ref{eq:logist}) with $r=4$ can be interpreted as being statistically equivalent to  a realization of a random variable with a distribution defined by $\varrho(z)$.

The Beta distribution
\begin{equation} \mathcal{B}(z,\alpha,\beta)=\frac{z^{\alpha-1}(1-z)^{\beta-1}}{\mathbb{B}(\alpha,\beta)} \end{equation}
constitutes a family of continuous probability distributions. It is defined on the (sample space) interval $[0,1]$  and  its shape is determined by the Beta function $\mathbb{B}(\alpha,\beta)$  with two positive parameters $\alpha$ and $\beta$. 
 For $\alpha=\beta=0.5$, the Beta function $\mathbb{B}(0.5,0.5)=\pi$;   consequently, the Beta distribution $\mathcal{B}(0.5/0.5)$ and the natural invariant density of the Logistic map (\ref{eq:logist}) equal as $\mathcal{B}(z,0.5,0.5)=\varrho(z)$. In other words,  chaotic sequences of the Logistic map with $r=4$ and random sequences obtained from realizations of the Beta distribution with $\alpha=\beta=0.5$ have the same density and are statistically equivalent~\cite{diak96}. This equivalence opens up comparing the effect of random and chaotic sequences on the performance of PSO.

For  studying the effect of such a statistical equivalence also for other shapes, we consider 5 more chaotic maps. Three of them have natural invariant densities which  match frequently used random distributions. For the other two maps there is no direct random equivalent.  
The three maps matching are the Chebyshev map
\begin{equation}z(k+1)=\cos{(a \arccos{(z(k))} } \label{eq:cheb} \end{equation}
(chaotic for $a=6$), the Weierstrass map
\begin{equation}z(k+1)=\sum_{i=0}^N a^i \cos{(b^i\pi z(k))} \label{eq:weier} \end{equation}
(chaotic for $a=0.999$,
$b=101$ and $N=100$), and the Tent map
\begin{equation}z(k+1)=r \: \min{(z,1-z)},   \label{eq:tent} \end{equation}
which is chaotic for $r=2$. The Tent map \eqref{eq:tent} gives sequences which are statistically equivalent to a uniform distribution $\mathcal{U}(0,1)$ (and $\mathcal{B}(1/1)$), the Chebyshev map is statistically equivalent to the Logistic map with $r=4$ and thus to $\mathcal{B}(0.5/0.5)$,  and the Weierstrass map \eqref{eq:weier} is at least approximately equivalent to a normal distribution $\mathcal{N}(\mu/\sigma)$~\cite{law14} (and $\mathcal{B}(\alpha/\beta)$ for large $\alpha=\beta$).
The two maps remaining are the Cubic map
\begin{equation}z(k+1)=rz(k)(1-z(k)^2) \label{eq:cubiclogist} \end{equation}
 exhibiting chaotic behaviour for $r=2.62$ and the Bellows map
\begin{equation}z(k+1)=\frac{rz(k)}{1+z(k)^6}, \label{eq:bellows} \end{equation}
which is chaotic for $r=2$. 
As we want to use these maps analogously to the previously described Logistic map for providing chaotic sequences with certain statistical properties, the trajectories  are re-scaled to the (sample space) interval $[0,1]$.

\begin{figure}[htb]
\begin{center}
\includegraphics[trim = 35mm 90mm 42mm 100mm,clip, width=6.2cm, height=5.2cm]{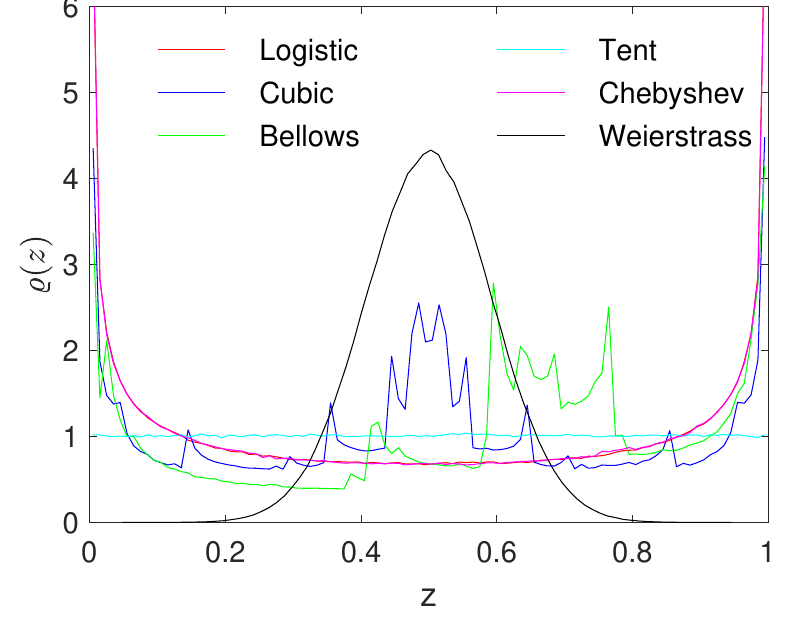}
\includegraphics[trim = 35mm 90mm 42mm 100mm,clip, width=6.2cm, height=5.2cm]{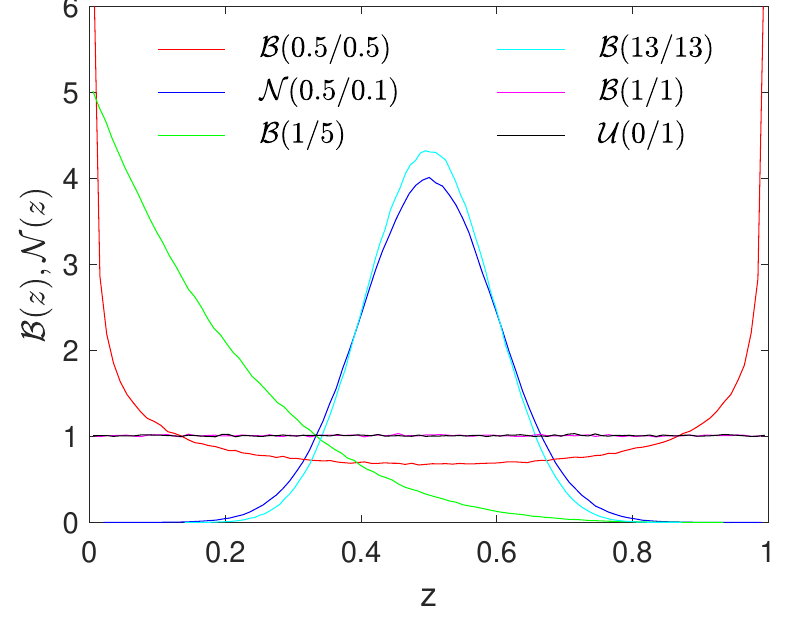}

(a) \hspace{4cm} (b)
\caption{Distribution of sequences used in the study. (a) Natural invariant densities of chaotic maps. (b) Probability density functions of random variables. }
\label{fig_chaotic}
\end{center}
\end{figure}

 To have an illustration of the differences in the statistical properties of the sequences, Fig.~\ref{fig_chaotic}(a) shows numerically calculated approximations of the invariant densities for all 6 chaotic maps considered in the study. The results are averages over 10.000 different initial states and 400 iterations each. We see that the Logistic map and the Chebyshev map have a bathtub-shaped invariant density, while for  the Weierstrass map we find a bell-shaped curve. The Tent map has a constant distribution over the sample space interval $[0,1]$. By contrast, 
  the Cubic map and the Bellows map have distributions with several peaks, where for the Cubic map we find symmetry on the sample space interval $[0,1]$, while for the Bellows map there is an asymmetric distribution with most peaks for values in the interval $[0.6,0.8]$.

The random sequences used in this study are generated with a Mersenne Twister PRNG~\cite{mat98}. Starting with a seed value such a generator produces a sequence as a realization of (or at least an imitation of) a random variable with prescribed probability distribution. Fig.~\ref{fig_chaotic}(b) shows the probability density functions of $\mathcal{B}(0.5/0.5)$, $\mathcal{B}(1/5)$, $\mathcal{B}(1/1)$, $\mathcal{B}(13/13)$, $\mathcal{U}(0/1)$ and $\mathcal{N}(0.5/0.1)$ and thus provides a comparison of the distributions for the random sequences considered here. The functions of      
$\mathcal{B}(0.5/0.5)$, the Logistic map and the Chebyshev map, but also the functions of $\mathcal{B}(1/1)$, $\mathcal{U}(0/1)$ and the Tent map, are the same. Also the functions of $\mathcal{N}(0.5/0.1)$, $\mathcal{B}(13/13)$, and the Weierstrass map resemble each other.  

\begin{figure}[htb]
\begin{center}
\includegraphics[trim = 35mm 90mm 42mm 100mm,clip, width=6.2cm, height=5.2cm]{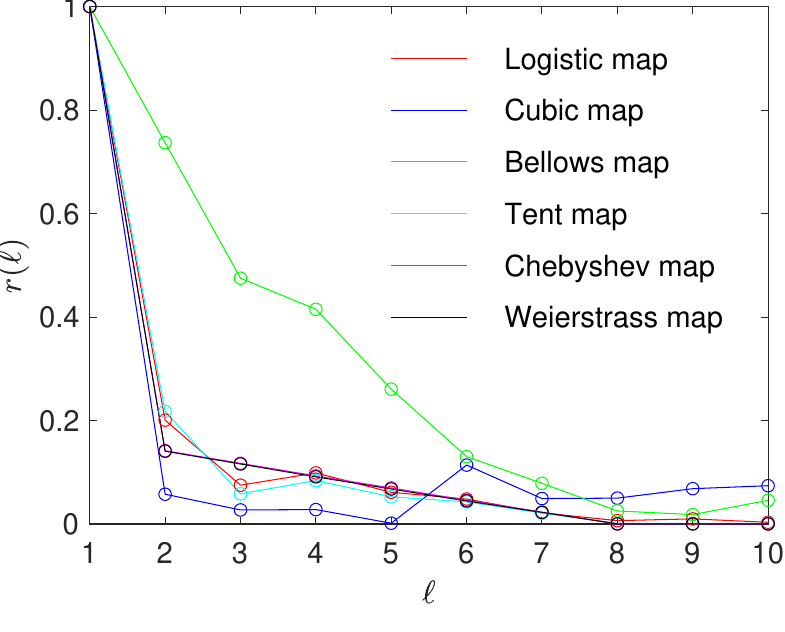}
\includegraphics[trim = 35mm 90mm 42mm 100mm,clip, width=6.2cm, height=5.2cm]{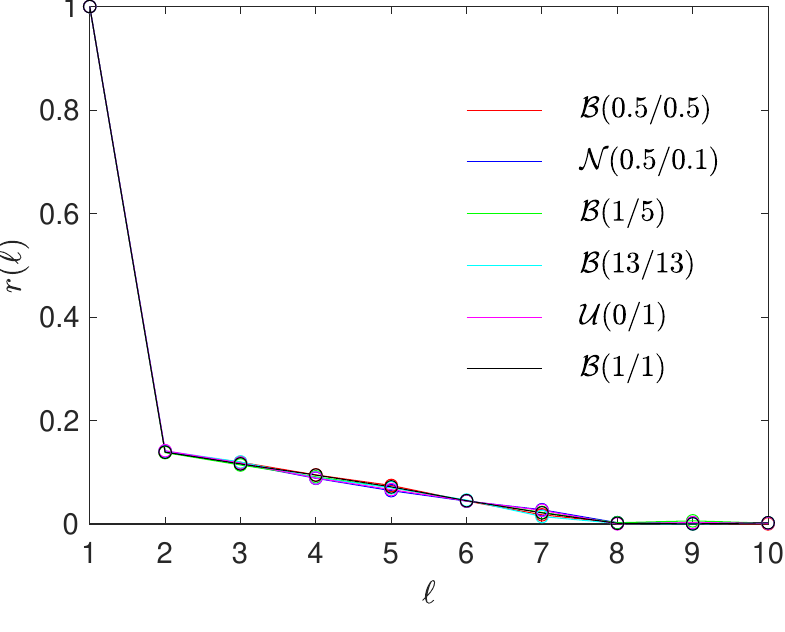}

(a) \hspace{4cm} (b)
\caption{Estimation $r(\ell)$, Equation \eqref{eq:corr}, of the autocorrelation of the sequences with time lag $\ell$.    (a) Chaotic maps. (b) Random variables. }
\label{fig_corr}
\end{center}
\end{figure}

Apart from differences in the underlying distributions as discussed so far, also the effects of short-term correlations are of interest. We consider two ways to account for these effects, the Lyapunov exponent of the chaotic maps and the autocorrelation between successive values. The Lyapunov exponent of a dynamical system quantifies the degree of separation between sequences starting infinitesimally close to each other, see e.g.~\cite{ott93},  p. 129.  Thus, the larger the Lyapunov exponent is the faster sequences diverge on average, which is one way to  capture dependencies between successive values in a sequence.  
Tab.~\ref{tab:map} gives the Lyapunov exponent $\lambda$ for the chaotic maps considered. Again,  the results are averages over 10.000 different initial states and 400 iterations each. We see that the Logistic, the Cubic and the Tent map have almost the same value of $\lambda$, while for the Bellows map, it is significantly smaller and for the Chebyshev map it is considerably larger. For the Weierstrass map, the Lyapunov exponent is dramatically larger.

\begin{table}
\caption{Lyapunov exponents and AUC of the autocorrelation of the chaotic maps considered.}
\begin{tabular}{llll} \hline Map & Equation & Lyapunov exponent $\lambda$ & AUC \\ \hline
    Logistic map &  \eqref{eq:logist} & $0.6916$ &  1.0351\\ 
    Chebyshev map &  \eqref{eq:cheb} & $1.7918$ & 1.0022 \\
    Weierstrass map &  \eqref{eq:weier} & $452.6356$ &  0.9941 \\
       Tent map &  \eqref{eq:tent} & $0.6931$ & 0.9900 \\
     Cubic map &  \eqref{eq:cubiclogist} & $0.6915$ & 0.9429 \\
    Bellows map &  \eqref{eq:bellows} & $0.3761$ & 2.6926\\
 
 \hline
    \end{tabular}
\label{tab:map}
     \end{table}

 Finally, for  having another measure for differences in short-term correlations between successive values, we  estimate the  autocorrelation of a sequence with time lag $\ell$ by 
\begin{equation} r(\ell)=\frac{\sum_{t=0}^{T-\ell}(z(t+1)-\bar{z})(z(t+\ell)-\bar{z})}{\sum_{t=0}^{T-1} (z(t+1)-\bar{z})^2} \label{eq:corr}\end{equation}
with $\bar{z}=\frac{1}{T} \sum_{t=0}^{T-1} z(t+1)$, see e.g~\cite{kan97}. The estimation of the autocorrelation $r(\ell)$ compares sections of the sequences and shows how they correlate, see Fig.~\ref{fig_corr} for results of the same chaotic and random sequences as in Fig. \ref{fig_chaotic}. 
The results are obtained by averaging 10.000 samples from different initial states or seeds and $T=400$. 
For $\ell=1$, we compare the same section, thus obtaining $r(1)=1$.   For $\ell>1$, the correlation decays rapidly, but the rate of decay differs between the chaotic sequences, but not between the random sequences obtained by the PRNG. For chaotic sequences, the correlation decays slowest for the Bellows map, and fastest for the Cubic map, with the Logistic, the Tent, the Chebyshev  and the Weierstrass map between them. By comparing with Fig. \ref{fig_corr}(b), we see that the short-term correlations of the PRNGs are almost identical to the  Chebyshev and the Weierstrass map, while for the Logistic and the Tent map, we find generally a clear similarity, but for $\ell=2$ and $\ell=3$  there are also small but statistically significant differences. The Cubic and the Bellows map yield    correlation decays markedly different from these maps, and also from PRNGs. To have a quantification    of the rate of decay, we use the area under the curve (AUC), which is the (numerical) integration of the area bounded by $r(\ell)$  for $1 \leq \ell \leq 10$. The AUC measures the rate of correlation decay, with a rapid decay yielding smaller values, while a slower decay gives larger values. See Tab.~\ref{tab:map} for the AUC of the chaotic maps normalized by the AUC of the PRNGs (the AUC of the PRNGs is 0.9888, which means  the results after normalization are not very different from the original AUC). 

Comparing the results for the Lyapunov exponent and the autocorrelation we see that to some extend similar effects are captured. For instance, the low rate of divergence characterized by the low Lyapunov exponent $\lambda$ of the Bellows map is mirrored by the large AUC indicating a slow decay of the autocorrelation. Also, for the Logistic, the Cubic and the Tent map, $\lambda$ and AUC give consistent results. For the Chebyshev and the Weierstrass map this is not the case. Although their Lyapunov exponents are (much) larger, the AUC is comparable to the other maps (except the Bellows map). This underlines that the average divergence of nearby trajectories underlying a sequence and the autocorrelation between sections of a sequence are not exclusively accounting for the same effect.

 \section{Experimental setup}
 \subsection{PSO algorithm}
\textit{Particle swarm optimization} (PSO) is a generational population-based algorithm used for calculating optima of single- and multidimensional functions. A PSO particle holds 4 variables for each  generation $t$: its position $x(t) \in \mathbb R^D$ and velocity $v(t)\in \mathbb R^D$ in search space, its own best position (local best position) $p^l_t$ and the best position of any particle of the population that occurred during all generations (global best position) $p^g_t$. The particle movement is described by  
\begin{align}
x(t+1) &=x(t) + v(t+1) \label{eq:pos}\\
v(t+1) &=wv(t)+c_1r_1(p^l_t - x(t))+c_2r_2(p^g_t - x(t)) \label{eq:vel}.
\end{align}
The parameters  $w$, $c_1$ and $c_2$ are the inertial,  cognitive and social weights. The random variables $r_1,r_2$ are taken from the sequences produced by either the PRNGs or chaotic maps. 
At $t=0$, each particle $i$, $i=1,2,\ldots I$,  of  swarm size $I$ is initialized randomly in position $x(0)$ with velocity $v(0)$. 
The following steps are repeated for $t>0$ until some stop criterion is fulfilled:
\begin{itemize}
    \item[](1) update velocity for each particle by Equation \eqref{eq:vel}
    \item[](2) update position of each particle by Equation \eqref{eq:pos}
    \item[](3) evaluate fitness function for each particle
    \item[](4) set new local  and global bests, $p_t^l$ and $p^g_t$, if needed.
\end{itemize}
\noindent  In the simulation we take standard values of PSO parameters frequently used:
 \begin{itemize}
     \item[] $w = 0.79$ \hspace{2cm} $200$ generations per run
     \item[] $c_1 = 1.49$ \hspace{2cm} $4000$ runs per sample
     \item[] $c_2 = 1.49$ \hspace{2cm} $I=100$ individuals per swarm.
 \end{itemize}
\subsection{Test functions} \label{sec:test}
To compare the impact of chaotic and random sequences on PSO the following test functions $f(x)$ from the IEEE CEC 2013 test suite~\cite{li13} and a recent study on benchmarking~\cite{kud22} are employed:
\begin{itemize}
    \item 1 Equal Maxima (1 dimension)
    \item 2 Uneven Decreasing Maxima (1 dimension)
    \item 3 Himmelblau (2 dimension)
    \item 4 Six-Hump Camel Back (2 dimension)
    \item 5 Shubert (2 dimension)
    \item 6 Vincent (2 dimension)
    \item 7-9 Rastrigin (10, 20 and 30 dimensions)
    \item 10-12 Rosenbrook (10, 20 and 30 dimensions)
    \item 13-15 Sphere (10, 20 and 30 dimensions)
      \item 16-18 Ackley (10, 20 and 30 dimensions)
        \item 19-21 Griewank (10, 20 and 30 dimensions)
          \item 22-24 Penalized1 (10, 20 and 30 dimensions)
            \item 25-27 Penalized2 (10, 20 and 30 dimensions).
\end{itemize}
The test functions cover a fairly wide range of problem difficulty, with 
most of the functions being multi-modal, while the functions 10-12 (Rosenbrock) and 13-15 (Sphere) are uni-modal. Furthermore, while most of the functions are non-separable, three of the functions are separable (7-9, Rastragin, 13-15, Sphere, and 25-27, Penalized2).
\subsection{Distributions}
\label{sec:dist}
Random sequences with the following distributions (generated by PRNGs) are used: 
\begin{itemize}
    \item[] (a) $\quad \mathcal{B}(0.5/0.5)$ \hspace{2.1cm}  (d) $\quad \mathcal{U}(0/1)$
    \item[] (b) $\quad \mathcal{N}(0.5/0.1)$ \hspace{2cm}  (e) $\quad \mathcal{B}(1/1)$
    \item[] (c) $\quad \mathcal{B}(13/13)$ \hspace{2.3cm} (f) $\quad \mathcal{B}(1/5)$
      \end{itemize} 
Chaotic sequences from chaotic maps are taken according to Tab.~\ref{tab:map}.

\subsection{Performance evaluation}
The performance of PSO runs is quantified by the mean distance error MDE,  which is the mean of the absolute difference in search space between the found best fitness value and the global optimum of the function $f(x)$. 
 A good performance means a low mean distance of the found best result to the actual optimum of the fitness function.


\section{Results}

The results presented in this section rely upon analyzing performance data of the 27 test functions (see Sec. \ref{sec:test}) over 6 chaotic and  6 random sequences (see Sec. \ref{sec:dist}) for 4000 runs~\footnote{The raw data of these $27\times12\times 4000=1.296.000$ performance data points can be found at the data repository \url{https://github.com/HendrikRichterLeipzig/Random_Chaos_PSO}}.

\begin{figure}[htb]
\begin{center}
\includegraphics[trim = 25mm 90mm 32mm 90mm,clip, width=8.2cm, height=6.2cm]{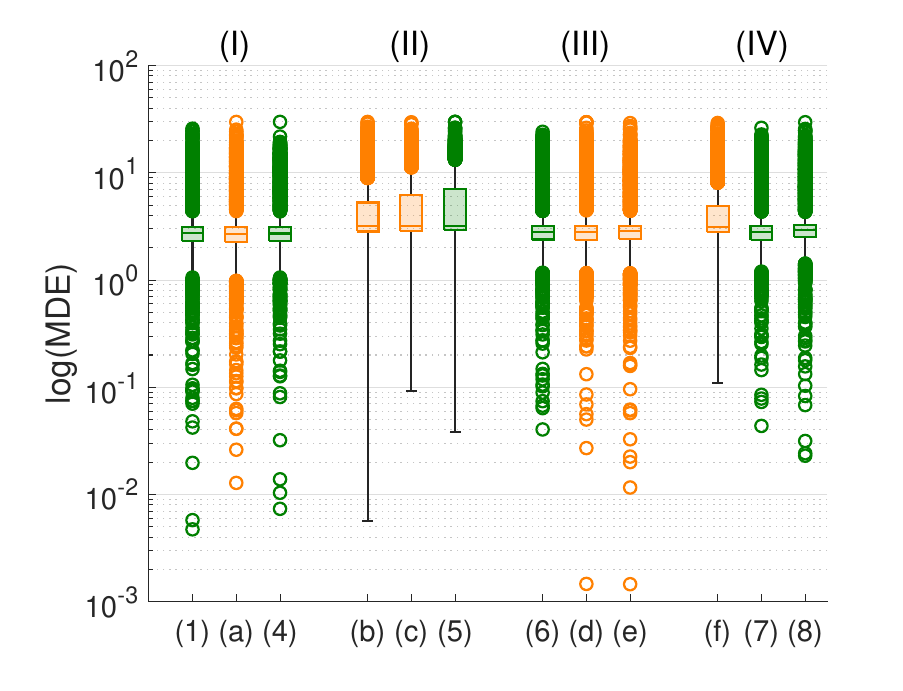}

(a)

\includegraphics[trim = 25mm 90mm 32mm 90mm,clip, width=8.2cm, height=6.2cm]{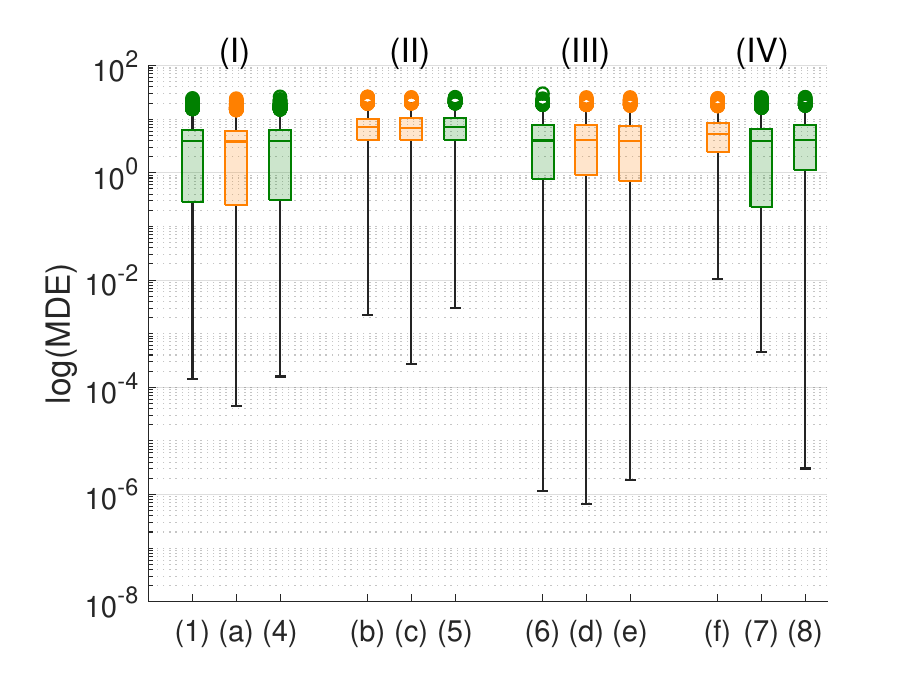}

(b)

\end{center}

\caption{Boxplots of PSP performance with green boxes indicating chaotic maps and orange boxes indicating random by PRNGs. (a)  Rosenbrock, $D=10$ (b) Penalized 1, $D=10$. }
\label{fig_box}
\end{figure}

A first step in analyzing the data is by displaying the performance measured by the MDE over 4000 PSO runs as boxplots. A boxplot gives a graphical representation of the
locality, spread and skewness groups of performance data through their quartiles. In addition, outliers differing significantly from the rest of the dataset are plotted above of below the whiskers of the boxplot.  Thus, a boxplot not only delivers the mean  of the result data~\footnote{This quantity can also be obtained from the data repository.}, but also quartiles and outliers, and gives a more nuanced evaluation of the data. 
In the boxplot of Fig. \ref{fig_box}, the performance for the 6 chaotic and 6 random sources of sequences are given. Green boxes indicate chaotic maps as specified by the equation numbers in Tab.~\ref{tab:map}, while orange boxes indicate PRNGs by the letters (a),(b),...,(f) as given in Sec.~\ref{sec:dist}. The results are grouped according to the shape of the distributions, with $\text{(I)}=\text{(1),(a),(4)}$ representing bathtub-shaped, $\text{(II)}=\text{(b),(c),(5)}$ bell-shaped, and $\text{(III)}=\text{(6),(d),(e)}$ constant distributions over the sample space interval $[0,1]$; $\text{(IV)}=\text{(f),(7),(8)}$ are the remaining distributions, which are different from each other.

We take two examples which are typical for the test functions considered. The first example is the $D=10$ dimensional Rosenbrock function, see Fig. \ref{fig_box}(a). We get for all sequences almost the same mean (with the exception of the $\mathcal{B}(1/5)$, see the boxplot (f) in the group (IV). In other words, if we were just to use the mean MDE over runs, we might conclude that roughly speaking all sequences (except $\mathcal{B}(1/5)$) perform almost equally. However, the results shown in Fig. \ref{fig_box}(a) are more subtle.  For instance, the bell-shaped distributions, group (II), there are no lower outliers indicating rare but occasionally occurring very good performances, while at the same time upper outlier indicating poor performance are also less frequent. Moreover, the upper quartile is much higher than for the other distributions (again except $\mathcal{B}(1/5)$), and there is a difference in the upper quartile for the three sources of bell-shaped distributions (group (II): $\mathcal{N}(0.5/0.1)$,  $\mathcal{B}(13/13)$, and Weierstrass map), while for the bathtub-shaped and constant distributions (group (I) and (III)) the quartiles are the same.

\begin{table*}[htb]
  \centering
   \caption{\small{Performance comparison matrix showing results for all 27 test functions  and all 12 distributions of chaotic and random sequences. For each cell the tuple W$[+,-,\thickapprox]$ (using the Wilcoxon rank-sum test)  and F$[+,-,\thickapprox]$ (using the Friedman test) indicates for $+$ that the sequences from the distribution in the row performs better, for $-$ that the sequences from the distribution in the column performs better, while for $\thickapprox$  there is no difference in performance. The upper triangle gives W$[+,-,\thickapprox]$ and lower triangle shows F$[+,-,\thickapprox]$. Note that the results transpose, that is, if the cell $(i,j)$ has the result $[+,-,\thickapprox]$, then the cell $(j,i)$ has $[-,+,\thickapprox]$.}    }
\scalebox{0.53}{
  \begin{tabular}{|c||c|c|c||c|c|c||c|c|c||c|c|c|}
  Group&(I)&(I)&(I)&(II)&(II)&(II)&(III)&(III)&(III)&(IV)&(IV)&(IV) \\
    \hline &
 Logistic, \eqref{eq:logist} & $\mathcal{B}(0.5/0.5)$, (a)  &  Chebyshev, \eqref{eq:cheb}  &  $\mathcal{N}(0.5/0.1)$, (b) & $\mathcal{B}(13/13)$, (c) & Weierstrass, \eqref{eq:weier} & Tent, \eqref{eq:tent} & $\mathcal{U}(0/1)$, (d) &  $\mathcal{B}(1/1)$, (e) &  $\mathcal{B}(1/5)$, (f) & Cubic, \eqref{eq:cubiclogist} & Bellows, \eqref{eq:bellows} \\ \hline
  Logistic, \eqref{eq:logist}& &W[6,5,16]&W[6,2,19]&W[14,5,8]&W[14,5,8]&W[14,4,9]&W[9,8,10]&W[7,8,12]&W[7,6,14]&W[10,9,8]&W[8,9,10]&W[16,2,9] \\ \hline
 $\mathcal{B}(0.5/0.5)$, (a)  & F[2,4,21] & & W[4,1,22] & W[12,8,7]&W[12,7,8]& W[13,6,8] &W[9,8,10]&W[9,8,10]&W[9,8,10]&W[12,9,6] &W[10,9,8]&W[17,1,9]  \\ \hline
 Chebyshev, \eqref{eq:cheb} &F[3,6,18]&F[0,4,23]&&W[7,13,7]&W[13,7,7]&W[13,5,9]&W[11,6,10]&W[7,8,12]&W[8,7,12]&W[11,9,7]&W[8,9,10]&W[17,1,9] \\ \hline
 $\mathcal{N}(0.5/0.1)$, (b)  &F[5,13,9]&F[8,12,7]&F[6,13,8]&&W[1,0,26]&W[8,3,16]&W[4,13,10]&W[5,16,6]&W[4,15,8]&W[4,12,11]&W[4,16,7]&W[6,9,12] \\ \hline
 $\mathcal{B}(13/13)$, (c) &F[5,12,10]&F[7,12,8]&F[9,13,5]&F[1,1,25]&&W[4,1,22]&W[5,12,10]&[4,15,8]&W[3,15,9]&W[4,13,10]&W[3,16,6]&W[9,9,9] \\ \hline
 Weierstrass, \eqref{eq:weier} &F[4,15,8]&F[6,13,8]&F[5,14,8]&F[3,8,16]&F[1,2,24]&&W[4,16,7]&W[3,16,8]&W[3,16,8]&W[4,15,8]&W[4,17,6]&W[7,12,8] \\ \hline
 Tent, \eqref{eq:tent} &F[6,10,11]&F[6,10,11]&F[6,12,9]&F[13,4,10]&F[12,5,10]&F[16,3,8]&&W[6,6,15]&W[4,4,19]&W[10,10,7]&W[2,15,10]&W[13,4,10] \\ \hline
 $\mathcal{U}(0/1)$, (d) &F[6,7,14]&F[7,9,11]&F[7,7,13]&F[16,3,8]&F[14,4,9]&F[16,3,8]&F[3,6,18]&&W[1,0,26]&W[10,8,9]&W[2,12,13]&W[14,4,9] \\ \hline
 $\mathcal{B}(1/1)$, (e) &F[6,7,14]&F[8,9,10]&F[7,10,10]&F[13,3,11]&F[13,3,11]&F[15,3,9]&F[3,3,21]&F[0,0,27]&&W[10,7,10]&W[2,1,14]&W[15,3,9] \\ \hline
 $\mathcal{B}(1/5)$, (f) &F[9,10,8]&F[9,12,6]&F[9,11,7]&F[12,4,11]&F[12,4,11]&F[15,4,8]&F[9,10,8]&F[7,11,9]&F[7,10,10]&&W[6,11,10]&W[9,10,8] \\ \hline
 Cubic, \eqref{eq:cubiclogist} &F[9,8,10]&F[9,8,10]&F[9,8,10]&F[15,3,9]&F[16,3,8]&F[17,4,6]&F[14,2,11]&F[11,2,14]&F[12,2,13]&F[11,6,10]&&W[17,4,6] \\ \hline
 Bellows, \eqref{eq:bellows} &F[2,16,9]&F[1,18,8]&F[1,17,9]&F[9,6,12]&F[9,8,10]&F[12,6,9]&F[5,13,9]&F[4,13,10]&F[3,15,9]&F[10,8,9]&F[4,17,6] &\\ \hline 
 \hline
 \parbox{1.5cm}{\centering chaotic \\ vs. random} &\parbox{1.5cm}{\centering W[58,38,66] \\ F[53,33,76]}&\parbox{1.5cm}{\centering W[58,31,73] \\ F[55,26,81]}&\parbox{1.5cm}{\centering W[53,42,67] \\ F[54,42,66]}&\parbox{1.5cm}{\centering W[34,68,60] \\ F[32,66,64]}&\parbox{1.5cm}{\centering W[33,65,64] \\ F[32,63,67]}&\parbox{1.5cm}{\centering W[20,72,70] \\ F[20,69,73]}&\parbox{1.5cm}{\centering W[53,38,71] \\ F[50,34,78]}&\parbox{1.5cm}{\centering W[54,39,69] \\ F[47,38,77]}&\parbox{1.5cm}{\centering W[50,36,76] \\ F[48,38,76]}&\parbox{1.5cm}{\centering W[58,56,48] \\ F[56,56,50]}&\parbox{1.5cm}{\centering W[75,27,60]  \\ F[74,24,64]}&\parbox{1.5cm}{\centering W[36,70,56] \\ F[36,68,58]}\\ \hline
    \hline
    \parbox{1.5cm}{\centering chaotic \\ vs. chaotic} &\parbox{1.5cm}{\centering W[53,25,57] \\ F[56,24,55]}&&\parbox{1.5cm}{\centering W[51,27,57] \\ F[54,27,54]}&&&\parbox{1.5cm}{\centering W[24,72,39] \\ F[22,74,39]}&\parbox{1.5cm}{\centering W[45,43,47] \\ F[43,44,48]}&&&&\parbox{1.5cm}{\centering W[67,26,42]  \\ F[66,26,43]}&\parbox{1.5cm}{\centering W[23,70,42] \\ F[24,70,41]}\\ \hline
    \hline
      \parbox{1.5cm}{\centering random \\ vs. random} &&\parbox{1.5cm}{\centering W[54,40,41] \\ F[53,40,42]}&&\parbox{1.5cm}{\centering W[22,55,58] \\ F[19,54,62]}&\parbox{1.5cm}{\centering W[18,56,61] \\ F[19,52,64]}&&&\parbox{1.5cm}{\centering W[50,26,59] \\ F[48,23,64]}&\parbox{1.5cm}{\centering W[48,24,63] \\ F[45,21,69]}&\parbox{1.5cm}{\centering W[49,40,46] \\ F[47,41,47]}&&\\ \hline
    \hline
  \end{tabular}}
  \label{tab:wilcoxon}
  \end{table*}

For the second example, the $D=10$ Penalized1 function, the results offer a slightly different evaluation. Here, we find different mean values for the distribution groups (I), (II) and (III) with sequences from the bell-shaped distribution performing slightly inferior as compared to bathtub-shaped and constant. Also, the lower whiskers indicating non-outlier minima are different, both among the groups and within the groups.  However, as for the 10D Rosenbrock function, Fig. \ref{fig_box}(a), for each distribution group the mean is the same and the quartile rather similar. This observation is particularly confirmed by the distribution group (IV), which are three unequal distributions  ($\mathcal{B}(1/5)$, Cubic map and Bellows map). For sequences from these distributions, we find differing means and differing quartiles. 

If we take these results for the 10D Rosenbrock and the 10D Penalized1 function in Fig. \ref{fig_box} as a starting point for formulating assumptions about the effect of chaotic and random sequences on PSO performance, we may note the following. Depending on the test function, different sequences used for driving a PSO algorithm can yield different or same performance as measured by averaging over  (a sufficiently large number of) runs.  Even if we obtain the same average performance, a more detailed analysis shows differences in quartiles and/or outliers for varying sequences. 
  In particular,  if we compare chaotic and random sequences with the focus of differences and equality (or at least similarities) in distribution, we notice
that generally significant differences occur when the underlying distributions differ. 
Furthermore, no real differences in performance are found between chaotic and random sequences if the distributions agree. 
However, even for the same distribution, 
some differences can be observed between sequences, namely in the quartile groups, the lower whiskers and in the outliers. This is an indication that while equivalence in performance corresponds to a general match in distribution,  the subtle differences in performance could be traced to other factors, for instance differences in short-term correlations. Next, we further formalize our analysis and consider two nonparametric tests.   We employ the two sided Wilcoxon rank-sum test which evaluates the null hypothesis that two
given data samples are samples from continuous distributions with
the same median and use a 0.05 significance level. Moreover, we also use the Friedman test, which is similar to  parametric two-way ANOVA, and choose a p-value of 0.05.

\begin{figure}[htb]
\begin{center}
\includegraphics[trim = 25mm 90mm 32mm 90mm,clip, width=9.2cm, height=6.8cm]{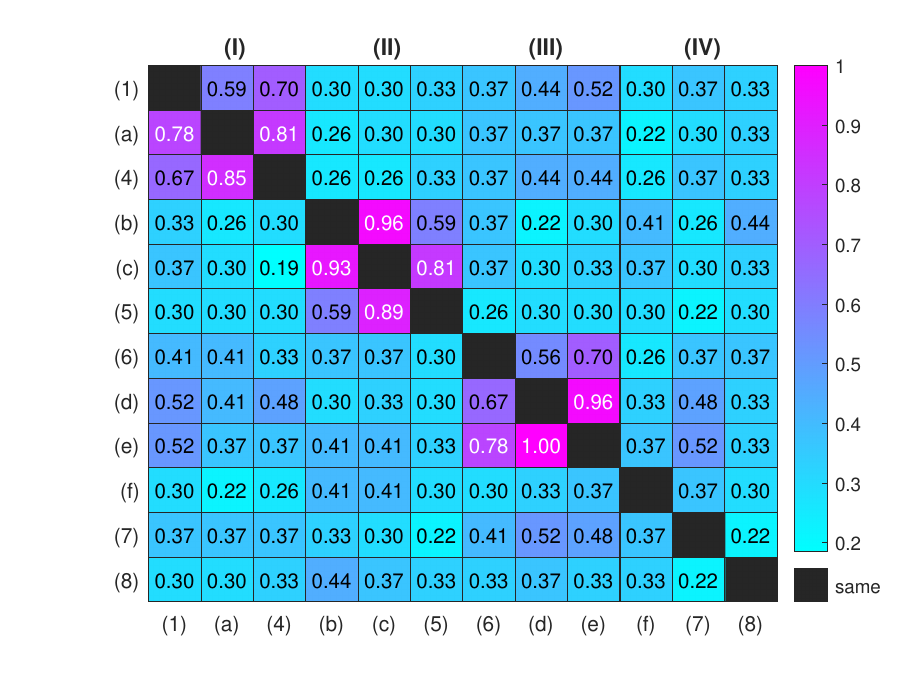}

\end{center}

\caption{Heat map of the fraction of indistinguishable results over all 27 test functions and 12 distributions.  i.e. $\frac{\thickapprox}{27}$ ; upper triangle, Wilcoxon test, lower triangle, Friedman test.}
\label{fig_heat}
\end{figure}
 Table \ref{tab:wilcoxon} gives for each comparison between different sequences a tuple with W$[+,-,\thickapprox]$ indicating results for the Wilcoxon test (and F$[+,-,\thickapprox]$ for the Friedman test), where
\begin{itemize}
    \item $+$  is the number of test functions where the PSO driven by the sequence in the \textit{row} performs better
    \item $-$  is the number of test functions where the PSO driven by the sequence in the \textit{column} performs better
    \item $\thickapprox $  is the number of test functions where the  PSO performances are not distinguishable by the Wilcoxon rank-sum test (or the Friedman test).
    \end{itemize}
The results confirm that performance differences primarily occur when the used sequences differ in their underlying probability distributions. For instance, compare the results for the Logistic map (which has a bathtub-shaped distribution) with the other two bathtub-shaped ($\mathcal{B}(0.5/0.5)$ and Chebyshev) and alternatively with the remaining 9 differently shaped distributions. We see that within the distribution group,  the Logistic map has for 16 and 19 (out of 27) test functions a performance  indistinguishable from the other two sources of sequences, while for 6 test functions, we have superior performance and for 5 and 2 test functions, performance is inferior.  Such a strong bias towards indistinguishable performance does not occur for any other sequence. There are, however, instances where the Logistic map is clearly better, with, for instance, 14 superior results (and 5 inferior and 8 equal results) compared to $\mathcal{N}(0.5/0.1)$ and  $\mathcal{B}(13/13)$. Yet such results do not necessarily imply that sequences from the Logistic map are generally better suited than other sequences for obtaining good PSO performance.    This can be seen, for instance, by comparing the Logistic map with sequences from distribution group (III), which has a constant distribution over the sample space interval $[0,1]$. Here, the number of indistinguishable   performances is lower (10, 12 and 14), but superior performances (9, 7, and 7) stand opposite to a similar number of inferior results (8, 8, and 6). In other words, the results differ more over test functions, but overall are still rather balanced. 

To emphasise the relationships between distribution and indistinguishable performances 
we visualize the result given in Tab. \ref{tab:wilcoxon} as a heatmap, see Fig. \ref{fig_heat}. As in the table the figure depicts the results for the Wilcoxon test in the upper triangle, while the lower triangle shows the results for the Friedman test. The heatmap shows darker, more purple or violet colors for larger fractions $\frac{\thickapprox}{27}$. We see that such higher fractions are obtained almost exclusively for results within a distribution group for the groups (I), (II) and (III). This indicates that indistinguishable results  primary occur if the distribution is the same (or very similar). Further note that although the exact values of the triple $[+,-,\thickapprox]$ slightly vary between the Wilcoxon test and the Friedman test, both tests give consistent results. This is also visible for the highest fractions of   indistinguishable results, which are obtained for comparing random sequences with the same distribution but different distribution function, that is for comparing  $\mathcal{N}(0.5/0.1)$ to  $\mathcal{B}(13/13)$  as well as comparing $\mathcal{U}(0/1)$ to  $\mathcal{B}(1/1)$. Similar high amounts of indistinguishable results, but a occasional better or worse performance can also be seen for re-runs with the same distribution.  These results are also supported by considering the performance in group (IV), which are for sequences with distributions  from $\mathcal{B}(1/5)$, Cubic map and Bellows map, and which are different from each other. We obtain a low level of indistinguishable results.  Moreover, while  $\mathcal{B}(1/5)$ and the Bellows maps perform rather similar (with 9 superior versus 10 inferior results), performance differences between   $\mathcal{B}(1/5)$ and the Cubic map, but also between the Cubic map and the Bellows map, are much stronger. 

Another interesting comparison is the overall performance of chaotic sequences vs. random sequences, see the summation in the last rows of Tab. \ref{tab:wilcoxon}, and relate these results to the overall performance of chaotic vs. chaotic and random vs. random. With the experimental setup of this paper, comparing chaotic vs. random can done for each sequence as we examine with respect to 6 different sources of random sequences (if the sequence is chaotic) or to 6 different sources of chaotic sequences (if the sequence is random), which gives $6 \times 27=162$ samples. Comparing chaotic vs. chaotic (or random vs. random) can be done for only 5 other sources and thus gives $5 \times 27=135$ samples. With this in mind, 
take, as an example, again the Logistic map, \eqref{eq:logist}, in the first column of Tab. \ref{tab:wilcoxon}.  According to the Wilcoxon test and for all 27 test functions and the 6 different random sequences considered, sequences from the chaotic Logistic map perform  with 58 superior results, against 38 inferior and  66 indistinguishable. Compared to the other 5 chaotic sequences, we get 53 superior, 25 inferior and 57 indistinguishable results. In other words, whereas the percentage of indistinguishable results is rather similar for comparing the Logistic map to other random sequences or to other chaotic sequences ($66/162 \approx 0.40$ and $57/135 \approx 0.42$), the percentage of inferior results is more different ($38/162 \approx 0.23$ and $25/135 \approx 0.18$). This suggests  that sequences from the Logistic map perform better than random sequences, but even better against  other chaotic sequences. However, for another example of chaotic sequences, the Bellows map, \eqref{eq:bellows}, last column of Tab. \ref{tab:wilcoxon}, the results are rather reversed. 
Here, and again according to the Wilcoxon test, 36 superior, 70 inferior and 56 indistinguishable results have been obtained against random sequences, while against chaotic sequences, we have 23 superior, 70 inferior and 42 indistinguishable results. Put differently, chaotic sequences from the Bellows map perform poor against random sequences, but even poorer against other chaotic sequences. Again, as for the individual performance comparisons between sequences, the results for the Wilcoxon test and for the Friedman test are different in detail, but give overall consistent results.

\begin{figure}[tb]
\begin{center}
\includegraphics[trim = 25mm 90mm 32mm 90mm,clip, width=9.2cm, height=6.7cm]{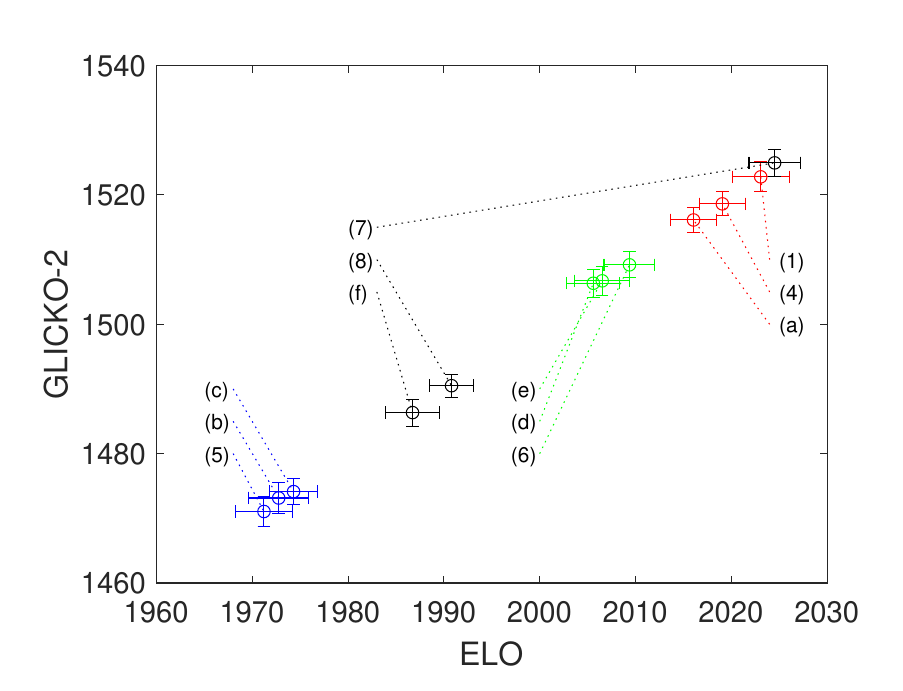}

(a)

\includegraphics[trim = 25mm 90mm 32mm 90mm,clip, width=9.2cm, height=6.7cm]{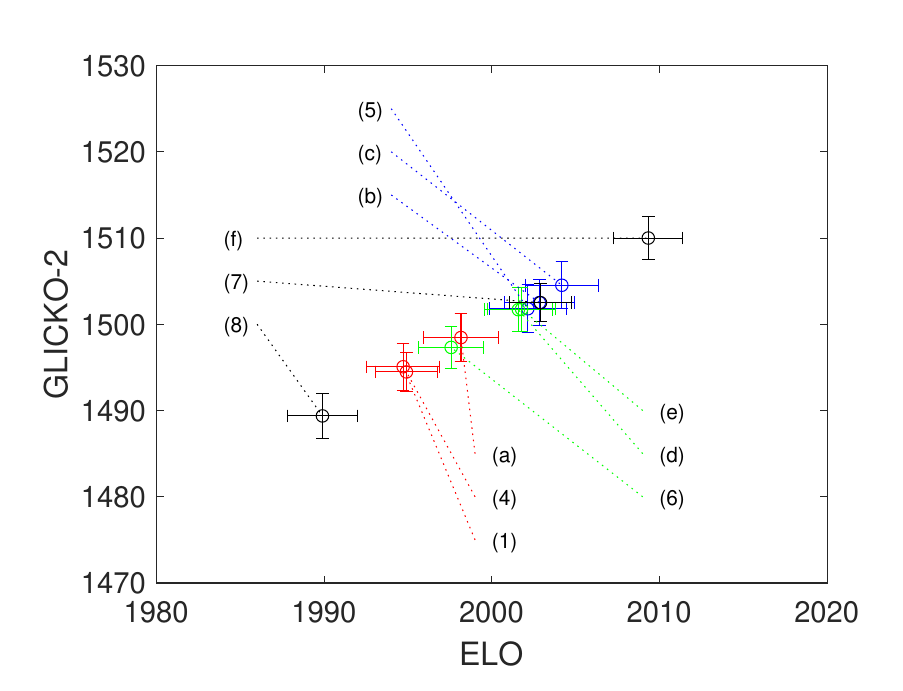}

(b)

\end{center}

\caption{Glicko-2 rating over Elo rating for comparing sequences. Colors indicates distribution groups; red:  (I)=(1),(4),(a), blue: (II)=(c),(b),(5), green: (III)=(e),(d),(6) black: (IV)=(7),(8),(f).  (a)  All 27 test functions according to Sec. \ref{sec:test}. (b) Only the test functions $\{1,2,3,4,5,6,7,10,11,15\}$. }
\label{fig_elo}
\end{figure}

We finally take another look at the performance data and analyse these data with a ranking method based on chess rating systems. For comparing results of evolutionary computation algorithms, it has been shown that the Glicko-2 rating provides particularly  meaningful evaluations, but to a somewhat lesser degree also the Elo rating can be applied~\cite{kud22,vecek14a,vecek14b}. Fig. \ref{fig_elo} shows  for all 12 chaotic and random sequences the Glicko-2 rating over the Elo rating. According to the chess rating system, each source of sequences is considered a player competing with the other sequence-players and getting awarded with a rating depending on performance, with larger values of Glicko-2 and Elo indicating better performance. The rating is calculated based on 4000 runs partitioned into 80 games between each sequence and calculated as suggested by Veček et al.~\cite{vecek14a,vecek14b}. The Glicko-2 calculation uses for each player an initial rating of $R=1500$, an initial rating deviation $RD=350$,  and  an initial rating volatility $\sigma=0.06$, while for the Elo calculation the initial rating is $E=2000$  and the K-factor $K=8$. Both rating methods use a draw threshold of $\epsilon=0.01$.  In Fig. \ref{fig_elo}, the distribution groups (I)--(IV) are indicated by different colors, and the circles give the mean rating, while the error bars  give the $95\%$ confidence intervals over the 80 games.  Fig. \ref{fig_elo}(a) shows the ratings over all 27 test functions. It can be seen that the Glicko-2 rating and the Elo rating evaluate the performance of sequences similarly. Between both ratings an almost linear relationship can be observed. However, the confidence intervals for the Elo rating are generally larger than for the Glicko-2 rating.  Furthermore, we see that the ratings cluster according to the distribution groups, with the exception of group (IV), which contains distributions differing from each other. From the ratings it could be concluded that the bathtub-shaped distributions (Logistic and Chebyshev map as well as $\mathcal{B}(0.5/0.5)$, (1),(4),(a) denoted in red), but also the Cubic map ((7), denoted in black) perform best, while bell-shaped distributions perform weakest. However, such a ranking also depends on the test functions. To illustrate this facts, Fig. \ref{fig_elo}(b) gives the  Glicko-2 rating and the Elo rating evaluating the performance of sequences, but only over a subset of the test functions, namely the functions $\{1,2,3,4,5,6,7,10,11,15\}$
according to Sec. \ref{sec:test}. For such a selection of test functions, the ranking of 
bathtub-shaped distributions and bell-shaped distributions is reversed. However, the clustering of distributions with the same shape still  applies equally.

\section{Conclusions}

The analysis given in this paper shows that some chaotic sequences perform better than random, but there are also some random sequences which perform better than chaos, which agrees with previously reported result~\cite{ala09,chen18,gan13,tian19,plu14,zel22,kuang14,rong10,yang12,liu05}. In view of these results it does not appear plausible to assume the presence of general and systematic differences in performance between chaotic and random sequences irrespective and  independent of taking into account the underlying distribution. For random and chaotic sequences with the same distribution (Logistic map and $\mathcal{B}(0.5/0.5)$, Chebyshev map and $\mathcal{B}(0.5/0.5)$, Weierstrass map and $\mathcal{N}(0.5/0.1)$ and $\mathcal{B}(13/13)$, Tent map and $\mathcal{U}(0/1)$ and $\mathcal{B}(1/1)$), we get the smallest differences in performance, while for all other combinations of distributions, we get more different performances.  These results support the conclusion that the underlying distribution rather than the origin is the main influential factor   in PSO performance.
In this line of argument, it appears to be not particularly meaningful to call a nature-inspired metaheuristic search algorithm which uses sequences from chaotic maps a ``chaotic metaheuristic algorithm''. Much more relevant would be to indicate which underlying distribution the chaotic sequences used actually have. 

The discussion about the presence and absence of differences in PSO performance relies upon evaluating benchmark functions. Bearing in mind that  only a finite number of such functions can practically be taken into account in numerical experiments, a sensible question to ask is whether  another selection might imply dramatically different results. For comparing different metaheuristic search algorithms, such a shift in results depending on the selection of problems has recently been demonstrated~\cite{delser21}. In fact, the results given in Fig.  \ref{fig_elo}  show such a shift in performance depending on which selection of test functions is actually used.
The conclusions of this paper, however, are different and not affected by such possible shifts in performance. We do not aim at find definite proof that sequences from any particular distribution generally outperform sequences from other distributions. We merely show that there are no intrinsic differences between using chaotic or random sequences irrespective and independent   of taking into account the underlying distribution. If such differences were to exist, they should show generally, no matter what collection of benchmark functions has been taken as long as it is not completely biased, which the selection of benchmark function considered, see Sec. \ref{sec:test}, is arguably not. However, if the selection of test functions used in this paper is fairly representative, then the results nonetheless illustrate that some distributions might be more promising than others for obtaining certain good results. It is particularly salient that bathtub-shaped distributions perform  well, incidentally with only small differences between chaotic and random sequences.   As the Logistic map, which is frequently used as a source of chaotic sequences for metaheuristic search, has exactly this distribution, it might be a reason why some previous works have seen advantages for chaos. Equally remarkable is the rather poor performance of bell-shaped distributions.

A final remark is about performance differences for same distributions. Although our results show that the distribution is the main influential factor in PSO performance, there are smaller differences which should  be attributable to other factors such as  short-term correlations between successive values. As shown in Sec. \ref{sec:rand_chao}, for chaotic sequences, short-term correlations vary considerably, see Fig.~\ref{fig_chaotic}(a).  Successive values of a realization of a random variable should, in theory,  be completely uncorrelated. However, practical implementations of PRNGs are not completely free of such short-term correlations, but  they are small and mainly constant for varying random distributions (and the same type of PRNG), see Fig.~\ref{fig_chaotic}(b).  Thus, even for the same (bathtub-like) shape of distribution, we find for comparing two chaotic sequences of different origin (Logistic and Chebyshev) larger differences in performance than between random and chaos (Logistic and $\mathcal{B}(0.5/0.5)$ or Chebyshev and $\mathcal{B}(0.5/0.5)$). Moreover, low Lyapunov exponents and large AUCs, which indicate slow decay of the autocorrelation appear to indicate poorer performance, compare Logistic and Chebyshev. A more detailed study of the relationships between short-term correlations and performance might be an interesting topic for future work.

\section{Data and source code availability}

 Source code of the numerical experiments as well as the
 performance data can be found at the data repository \url{https://github.com/HendrikRichterLeipzig/Random_Chaos_PSO}

\bibliographystyle{IEEEtran}
{\small
\bibliography{example}}

\begin{thebibliography}{10}
\providecommand{\url}[1]{#1}
\csname url@samestyle\endcsname
\providecommand{\newblock}{\relax}
\providecommand{\bibinfo}[2]{#2}
\providecommand{\BIBentrySTDinterwordspacing}{\spaceskip=0pt\relax}
\providecommand{\BIBentryALTinterwordstretchfactor}{4}
\providecommand{\BIBentryALTinterwordspacing}{\spaceskip=\fontdimen2\font plus
\BIBentryALTinterwordstretchfactor\fontdimen3\font minus
  \fontdimen4\font\relax}
\providecommand{\BIBforeignlanguage}[2]{{%
\expandafter\ifx\csname l@#1\endcsname\relax
\typeout{** WARNING: IEEEtran.bst: No hyphenation pattern has been}%
\typeout{** loaded for the language `#1'. Using the pattern for}%
\typeout{** the default language instead.}%
\else
\language=\csname l@#1\endcsname
\fi
#2}}
\providecommand{\BIBdecl}{\relax}
\BIBdecl

\bibitem{mat98}
M.~Matsumoto and T.~Nishimura, ``Mersenne twister: a 623-dimensionally
  equidistributed uniform pseudo-random number generator,'' \emph{ACM
  Transactions on Modeling and Computer Simulation (TOMACS)}, vol.~8, no.~1,
  pp. 3--30, 1998.

\bibitem{lec12}
P.~L’Ecuyer, ``Random number generation,'' in \emph{Handbook of Computational
  Statistics}, J.~E. Gentle, W.~K. H{\"a}rdle, and Y.~Mori, Eds.\hskip 1em plus
  0.5em minus 0.4em\relax Berlin, Heidelberg: Springer, 2012, pp. 35--71.

\bibitem{capo03}
R.~Caponetto, L.~Fortuna, S.~Fazzino, and M.~G. Xibilia, ``Chaotic sequences to
  improve the performance of evolutionary algorithms,'' \emph{IEEE Trans.
  Evolut. Comp.}, vol.~7, pp. 289--304, 2003.

\bibitem{chen18}
K.~Chen, F.~Zhou, and A.~Liu, ``Chaotic dynamic weight particle swarm
  optimization for numerical function optimization,'' \emph{Knowledge-Based
  Systems}, vol. 139, pp. 23--40, 2018.

\bibitem{gag21}
I.~Gagnon, A.~April, and A.~Abran, ``An investigation of the effects of chaotic
  maps on the performance of metaheuristics,'' \emph{Engineering Reports},
  vol.~3, no.~8, p. e12369, 2021.

\bibitem{xu18}
X.~Xu, H.~Rong, M.~Trovati, M.~Liptrott, and N.~Bessis, ``{CS-PSO}: chaotic
  particle swarm optimization algorithm for solving combinatorial optimization
  problems,'' \emph{Soft Comput.}, vol.~22, no.~3, pp. 783--795, 2018.

\bibitem{ma19}
Z.~Ma, X.~Yuan, S.~Han, D.~Sun, and Y.~Ma, ``Improved chaotic particle swarm
  optimization algorithm with more symmetric distribution for numerical
  function optimization,'' \emph{Symmetry}, vol.~11, no.~7, p. 876, 2019.

\bibitem{gan13}
A.~H. Gandomi, G.~J. Yun, X.-S. Yang, and S.~Talatahari, ``Chaos-enhanced
  accelerated particle swarm optimization,'' \emph{Commun. Nonlinear Sci.
  Numer. Simul.}, vol.~18, no.~3, pp. 327--340, 2013.

\bibitem{tian19}
D.~Tian, X.~Zhao, and Z.~Shi, ``Chaotic particle swarm optimization with
  sigmoid-based acceleration coefficients for numerical function
  optimization,'' \emph{Swarm Evol. Comput.}, vol.~51, p. 100573, 2019.

\bibitem{plu13}
M.~Pluhacek, R.~Senkerik, D.~Davendra, Z.~K. Oplatkova, and I.~Zelinka, ``On
  the behavior and performance of chaos driven {PSO} algorithm with inertia
  weight,'' \emph{Comp. Math. Appl.}, vol.~66, no.~2, pp. 122--134, 2013.

\bibitem{plu14}
M.~Pluhacek, R.~Senkerik, and I.~Zelinka, ``Particle swarm optimization
  algorithm driven by multichaotic number generator,'' \emph{Soft Comput.},
  vol.~18, no.~4, pp. 631--639, 2014.

\bibitem{yang14}
D.~Yang, Z.~Liu, and J.~Zhou, ``Chaos optimization algorithms based on chaotic
  maps with different probability distribution and search speed for global
  optimization,'' \emph{Commun. Nonlinear Sci. Numer. Simul.}, vol.~19, pp.
  1229--1246, 2014.

\bibitem{zel22}
I.~Zelinka, Q.~B. Diep, V.~Snasel, S.~Das, G.~Innocenti, A.~Tesi, F.~Schoen,
  and N.~V. Kuznetsov, ``Impact of chaotic dynamics on the performance of
  metaheuristic optimization algorithms: An experimental analysis,''
  \emph{Information Sciences}, vol. 587, pp. 692--719, 2022.

\bibitem{ala09}
B.~Alatas, E.~Akin, and A.~B. Ozer, ``Chaos embedded particle swarm
  optimization algorithms,'' \emph{Chaos, Solitons \& Fractals}, vol.~40,
  no.~4, pp. 1715--1734, 2009.

\bibitem{kuang14}
F.~Kuang, Z.~Jin, W.~Xu, and S.~Zhang, ``A novel chaotic artificial bee colony
  algorithm based on tent map,'' in \emph{Proc. 2014 IEEE Congress on
  Evolutionary Computation (CEC)}.\hskip 1em plus 0.5em minus 0.4em\relax
  Pistacaway, NJ: IEEE, 2014, pp. 235--241.

\bibitem{rong10}
H.~Rong, ``Study of adaptive chaos embedded particle swarm optimization
  algorithm based on skew tent map,'' in \emph{Proc. 2010 International
  Conference on Intelligent Control and Information Processing}.\hskip 1em plus
  0.5em minus 0.4em\relax Pistacaway, NJ: IEEE, 2010, pp. 316--321.

\bibitem{yang12}
C.-H. Yang, S.-W. Tsai, L.-Y. Chuang, and C.-H. Yang, ``An improved particle
  swarm optimization with double-bottom chaotic maps for numerical
  optimization,'' \emph{Applied Mathematics and Computation}, vol. 219, no.~1,
  pp. 260--279, 2012.

\bibitem{liu05}
B.~Liu, L.~Wang, Y.-H. Jin, F.~Tang, and D.-X. Huang, ``Improved particle swarm
  optimization combined with chaos,'' \emph{Chaos, Solitons \& Fractals},
  vol.~25, no.~5, pp. 1261--1271, 2005.

\bibitem{noer23}
P.~M. N{\"o}renberg and H.~Richter, ``Do random and chaotic sequences really
  cause different {PSO} performance?'' in \emph{Proc. GECCO 2023
  Companion}.\hskip 1em plus 0.5em minus 0.4em\relax New York: ACM, 2023, pp.
  99--102.

\bibitem{ott93}
E.~Ott, \emph{Chaos in Dynamical Systems}.\hskip 1em plus 0.5em minus
  0.4em\relax Cambridge, UK: Cambridge University Press, 1993.

\bibitem{diak96}
F.~K. Diakonos and P.~Schmelcher, ``On the construction of one-dimensional
  iterative maps from their variant density: The dynamical route to the beta
  distribution,'' \emph{Phys. Lett. A}, vol. 211, pp. 199--203, 1996.

\bibitem{law14}
M.~Lawnik, ``The approximation of the normal distribution by means of chaotic
  expression,'' \emph{Journal of Physics: Conference Series}, vol. 490, p.
  012072, 2014.

\bibitem{kan97}
H.~Kantz and T.~Schreiber, \emph{Nonlinear Time Series Analysis}.\hskip 1em
  plus 0.5em minus 0.4em\relax Cambridge, UK: Cambridge University Press, 1997.

\bibitem{li13}
X.~Li, A.~Engelbrecht, and M.~G. Epitropakis, ``Benchmark functions for
  {CEC}’2013 special session and competition on niching methods for
  multimodal function optimization.'' Tech. Rep., 2013, rMIT University,
  Evolutionary Computation and Machine Learning Group, Australia.

\bibitem{kud22}
J.~Kudela, ``A critical problem in benchmarking and analysis of evolutionary
  computation methods,'' \emph{Nature Machine Intelligence}, vol.~4, no.~12,
  pp. 1--8, 2022.

\bibitem{vecek14a}
N.~Veček, M.~Mernik, and M.~Črepinšek, ``A chess rating system for
  evolutionary algorithms: A new method for the comparison and ranking of
  evolutionary algorithms,'' \emph{Information Sciences}, vol. 277, pp.
  656--679, 2014.

\bibitem{vecek14b}
N.~Veček, M.~Črepinšek, M.~Mernik, and D.~Hrnčič, ``A comparison between
  different chess rating systems for ranking evolutionary algorithms,'' in
  \emph{2014 Federated Conference on Computer Science and Information
  Systems.}\hskip 1em plus 0.5em minus 0.4em\relax Pistacaway, NJ: IEEE, 2014,
  pp. 511--518.

\bibitem{delser21}
J.~{Del Ser}, E.~Osaba, A.~D. Martinez, M.~N. Bilbao, J.~Poyatos, D.~Molina,
  and F.~Herrera, ``More is not always better: Insights from a massive
  comparison of meta-heuristic algorithms over real-parameter optimization
  problems,'' in \emph{2021 IEEE Symposium Series on Computational Intelligence
  (SSCI)}.\hskip 1em plus 0.5em minus 0.4em\relax Pistacaway, NJ: IEEE, 2021,
  pp. 1--7.

\end{thebibliography}

\end{document}